%% file: acl2020.tex
\documentclass[11pt,a4paper]{article}
\usepackage[hyperref]{acl2020}
\usepackage{graphicx}
\usepackage{multicol}
\usepackage{amsmath}
\usepackage{times}
\usepackage{latexsym}

\usepackage{microtype}

\aclfinalcopy 

\title{Hierarchical Multi Task Learning with Subword Contextual Embeddings for Languages with Rich Morphology}

\author{Arda Akdemir \\
  University of Tokyo \\
  \texttt{aakdemir@hgc.jp} \\\And
  Tetsuo Shibuya \\
  University of Tokyo \\
  \texttt{tshibuya@hgc.jp} \\ \And
  Tunga G\"{u}ng\"{o}r\\
    Bogazici University \\
  \texttt{gungort@boun.edu.tr} \\
  }

\date{}

\begin{document}
\maketitle
\begin{abstract}
Morphological information is important for many sequence labeling tasks in Natural Language Processing (NLP). Yet, existing approaches rely heavily on manual annotations or external software to capture this information. In this study, we propose using subword contextual embeddings to capture the morphological information for languages with rich morphology. In addition, we incorporate these embeddings in a hierarchical multi-task setting which is not employed before, to the best of our knowledge. Evaluated on Dependency Parsing (DEP) and Named Entity Recognition (NER) tasks, which are shown to benefit greatly from morphological information, our final model outperforms previous state-of-the-art models on both tasks for the Turkish language. Besides, we show a net improvement of 18.86\% and 4.61\% F-1 over the previously proposed multi-task learner in the same  setting~\cite{akdemir2019joint} for the DEP and the NER tasks, respectively. Empirical results for five different MTL settings show that incorporating subword contextual embeddings brings significant improvements for both tasks. In addition, we observed that multi-task learning consistently improves the performance of the DEP component. 
\end{abstract}

\section{Introduction}

Multi-task Learning (MTL)  and  Language Modeling (LM) have both seen remarkable breakthroughs in recent years. MTL is shown to boost the performance of high-level tasks by leveraging the information obtained in low level tasks~\citep{collobert2011natural,Hashimoto2017AJM} and preventing deep learning models from overfitting a single domain. Language Models trained on huge unlabeled datasets such as ELMo~\citep{peters2018deep} and BERT~\citep{devlin2019bert} are successfully applied to many downstream NLP tasks.

\begin{figure}[ht]
    \includegraphics[scale=0.28]{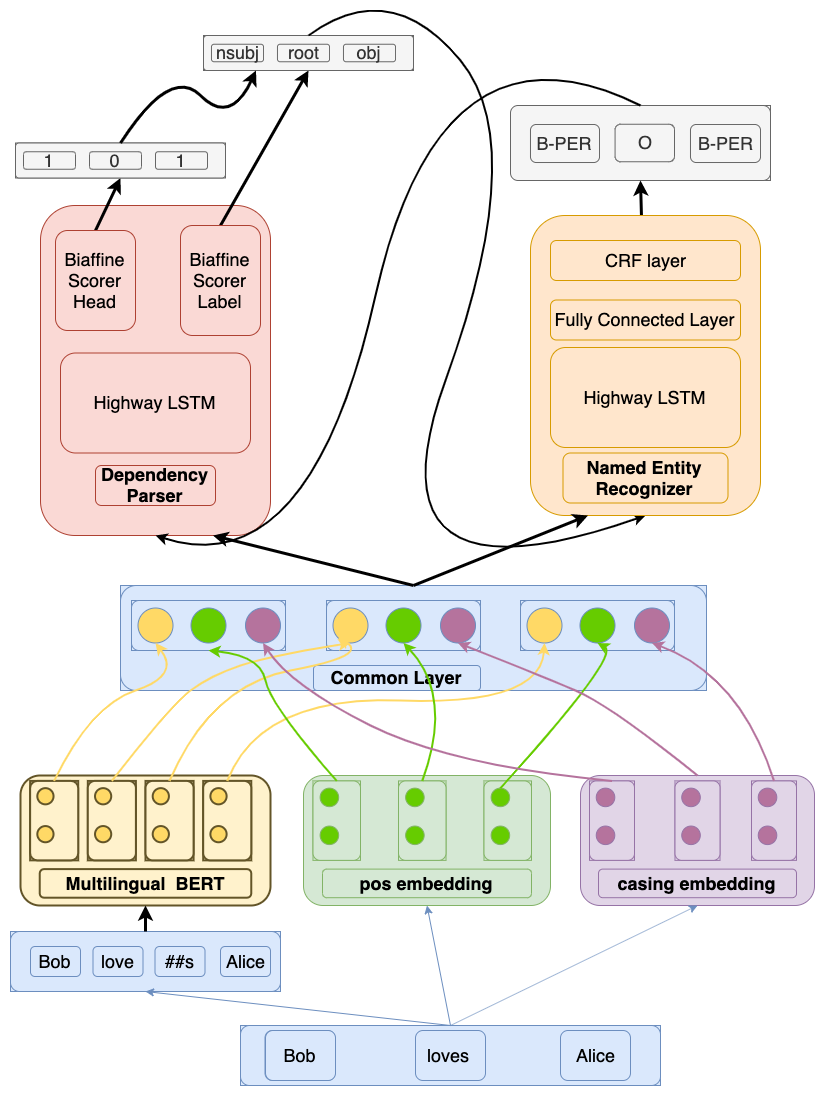}
    \caption{Overview of the proposed hierarchical multitask learning model using BERT representations. When DEP is considered as the low-level task we input the dependency prediction to the NER component (the outgoing arrow from the dependency label predictions). }
    \label{fig:pyJNERDEP}
\end{figure}

Rare and unknown words pose an important challenge for token-level vector representation based systems~\cite{luong2013better}. This is especially a major problem for languages with rich morphology like Turkish (Figures~\ref{rare_words} and~\ref{unk_words})\footnote{\label{note1} see Appendix~\ref{graphs} for details about how the figures are generated.}.
\begin{figure}[ht]
    \centering
    \includegraphics[scale=0.5]{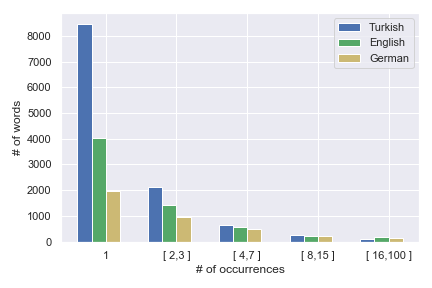}
    \caption{Word occurrences for Turkish, English and German. Rare words are much more common for Turkish.}
    \label{rare_words}
\end{figure}Linguistically motivated attempts focus on capturing the morphological information by dividing the tokens into sub-word  lexical units and use the vector representations of these units to  represent a token~\cite{luong2013better}. Early approaches mostly use character n-gram features~\cite{nadeau2007survey} or  morphological analyses~\cite{eryiugit2008dependency,gungor2018improving}. Recent works make use of character n-gram embeddings to capture the subword information~\citep{bojanowski2017enriching,lample2016neural}.
`

\begin{table*}[htbp]
    \centering
    \begin{tabular}{c|c c}
    \hline
       \textbf{Original Text }& \textbf{True Segmentation} & \textbf{BERT representation}\\\hline
         geliyordum & gel+iyor+du+m & gel+\#\#iy+\#\#ord+\#\#um\\\hline
         altında & alt+ı+nda& alt+\#\#ı+\#\#nda\\\hline

    \end{tabular}
    \caption{Overview of the sub-word tokens generated by BERT compared to the ideal segmentations. `geliyordum' means `I was coming' and `altında'  means `under (something)'.}
    \label{subword-example}
\end{table*}
In languages with a rich morphology like Turkish, words with 3-4 suffixes are quite common and important information is included in these morphological units rather than in the syntax~\citep{gungor2018improving}. For instance, the Turkish word `geliyordum', which means `I was coming', contains three suffixes appended to the root `gel' (to come): `(i)yor' denotes continuous tense, `du' denotes past tense and `m' denotes first person singular. Considering that these suffixes can be attached to any verb, a simple combinatorial calculation reveals that there would be many rare words in any corpus for such a language. Thus, it is very difficult to obtain good vector representations for words containing many suffixes. 
\begin{figure}[ht]
    \centering
    \includegraphics[scale=0.5]{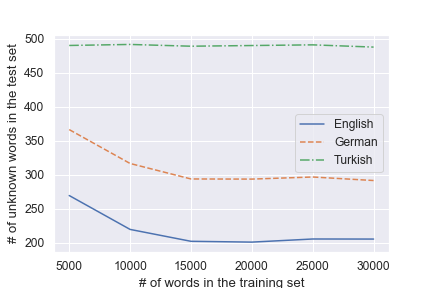}
    \caption{Number of unknown words in the test set for increasing sizes of the training set. }
    \label{unk_words}
\end{figure}

The importance of using morphological features and morpheme embeddings for Turkish language has been shown in various studies. Specifically, \citet{gungor2018improving,yeniterzi2011exploiting,demir2014improving} showed significant performance improvements for NER by using either hand-crafted morphological features or by using morphological embeddings. \citet{eryiugit2008dependency} obtained   significant gains for DEP by using a lexical-based rather than token-based approach where the lexical units are obtained using a morphological analyzer (see Appendix~\ref{related} for details about each model mentioned above). Yet, all models require either a manually annotated dataset for morphological analysis or an external analyzer for annotation.
Combining the above observations, we propose a new hierarchical MTL framework that uses subword contextual embeddings shared by both task-specific components. We claim that contextual subword representations can be a good replacement for morphological analyses for languages with rich morphology in an MTL setting. In this way, we extend the previous work on using contextual subword embeddings on sequence labeling~\citep{heinzerling2019sequence,che2018towards} and using contextual subword embeddings with multi-task learning~\citep{liu2019multi}.

Unlike~\citet{liu2019multi} who propose a flat model sharing only a common layer, we follow \citet{Hashimoto2017AJM} and implement a hierarchical MTL framework. The overall architecture is given in Figure~\ref{fig:pyJNERDEP}. The architecture consists of two task-specific components that share a common layer containing three types of embeddings. In the hierarchical multi-task settings, the output of the low-level task component is fed into the high-level task, i.e., the arrows coming out of dependency label predictions and named entity label predictions in Figure~\ref{fig:pyJNERDEP}. Only one of these arrows exists in any single setting.

To test our claim, we pick two NLP tasks for which morphological information is shown to be critical~\citep{eryiugit2008dependency,gungor2018improving}: Dependency Parsing and Named Entity Recognition. Previously, \citet{che2018towards} proposed a similar dependency parser which incorporates word-level ELMo embeddings~\citep{peters2018deep} and obtained the highest LAS score (overall and for the Turkish language) on the `CoNLL 2018 Shared Task'.  Our final model evaluated on the commonly used benchmarks outperform previous models on both tasks for Turkish language by using minimal features.  Our main contributions can be listed as follows:

\begin{itemize}
    
    \item We propose a new hierarchical multitask framework incorporating subword contextual representations of BERT~\citep{devlin2019bert}.
    
\item We outperfom the previous state-of-the-art models on both tasks for Turkish  language by 1) using subword contextual  embeddings, 2) implementing more sophisticated neural architectures, and 3) incorporating multi-task learning.

    \item We bring a net improvement over the  previous work~\citep{akdemir2019joint} on multitask learning of DEP and NER for Turkish by 18.86\% LAS and 4.61\% F-1, respectively.

    \item We experiment with  five different multitask learning settings and compare their  performances.
\end{itemize}

\section{Methodology}

\subsection{Contextual Subword Embeddings}

Representing words with pretrained and fixed vectors learned over huge unlabeled datasets~\citep{mikolov2013distributed} has been a major breakthrough in NLP research. However, word-level vector representations are not capable of capturing the sub-word level information which is important for many sequence labeling tasks such as NER, part-of-speech (POS) tagging,  and DEP~\citep{luong2013better}. For example,  having the  -(a)tion suffix often denotes that the word is a noun and a POS tagger can  exploit this  idea. 


Another main drawback of these approaches, and of any non-contextual vector representations is that they are independent of the context once they are learned.  For example, the word `bass' in `I like eating bass' and `I play bass guitar'  would be represented with the same vector even though the first one refers to a fish. Recent work on contextual representations overcome this by using a Long Short Term Memory (LSTM)~\cite{hochreiter1997long} or  a Transformer model~\cite{vaswani2017attention} to condition the output for a given word to its surrounding context.

In this study, we use BERT~\citep{devlin2019bert} which is a sub-word level pretrained Transformer-based language model for multiple languages. We claim that the information retained in the morphological units of a Turkish word can be captured through the contextual sub-word representations generated by BERT. Table~\ref{subword-example} shows some example sub-word tokens used by BERT. For the word `altında' (under) we see a perfect alignment between the true segmentation and the  BERT tokenization (see Appendix~\ref{berttoks} for how we obtain and map the BERT tokens). Even though BERT tokens do not perfectly align  with the true lexical units for many examples, we claim that the representations are still relevant as they are dependent on their surrounding context. 

\subsection{Common Layer}

Common layer, shared by all task-specific components, is the concatenation of three vectors. For each word $w_j$ in an input sentence, the output $o_j = v_j^{bert}\oplus v_j^{cas} \oplus v_j^{pos}$, where $v_j^{bert}$, $v_j^{cas}$, and $v_j^{pos}$ correspond to BERT embeddings, casing embeddings, and POS tag embeddings, respectively. To obtain $v_j^{bert}$,  we first tokenize the word into its BERT subtokens. Then for each subtoken we get the mean of hidden outputs of the final four layers of BERT following the suggestions of ~\citet{devlin2019bert}. Finally, we take the average over all subtokens to get the final representation. Lower casing is a common preprocessing step when using word embeddings to reduce the vocabulary size. To retain the casing information which is shown to be useful for NER task~\citep{heinzerling2019sequence,gungor2018improving} we use casing embeddings with five categories. An example to each category are as follows : `Title', `ALLCAPS', `lower' , `contains'apostrophe'  and `1234' (numeric). For POS embeddings we used the XPOS, Turkish language specific POS tags defined in the Universal Dependency format~\cite{zeman-EtAl:2018:K18-2}. XPOS is chosen primarily because both datasets for NER and DEP are already annotated for this feature.

\subsection{Dependency Parser}

\begin{figure}[ht]
    \centering
    \includegraphics[scale=0.25
    ]{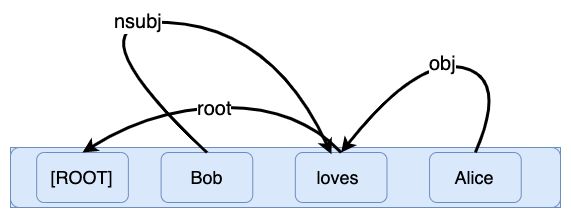}
    \caption{ We use two modules to generate scores for arcs (directed edges) and labels on the graph separately.}
    \label{depexp}
\end{figure}

We improved the previously proposed dependency parser for joint learning of the DEP and NER tasks (see Appendix~\ref{related} for details about the previous joint learner). The improved architecture (left part of Figure~\ref{fig:pyJNERDEP}) is heavily influenced by the graph-based parsers proposed by~\citet{qi2018universal,dozat2018simpler}. The parsers consist of two modules to generate scores for arcs and labels on the dependency graph shown in Figure~\ref{depexp}.

We first follow~\citet{qi2018universal} and replace the LSTM layers with a more sophisticated Highway-LSTM architecture (HLSTM)~\cite{srivastava2015highway}.  HLSTM learns adjustable parameters for controlling the information flow between LSTM layers (see Appendix~\ref{hlstm} for more details about Highway LSTM). Given a sequence of $n$ embeddings $\{o_t:1\leq t\leq n\}$,$$h_t = HLSTM^{dep} (o_1,...,o_n;t)$$
where $h_t$ is the hidden representation for $o_t$. \citet{dozat2018simpler} make a key observation that the vector representation of a word should be different when it is considered as a head or a dependent. Following this idea, these learned representations are fed into two Fully Connected layers ($FC_{head}^{edge}$ and $FC_{dep}^{edge}$) to generate representations for each word as a head and a dependent. In total we get four representations for each $h_t$. Next we apply deep-biaffine transformation to these vectors to generate scores for each arc and each label for each pair $i,j$.\begin{align*}
    v_i^{he}, v_j^{de} &= FC_{head}^{edge}(h_i), FC_{dep}^{edge}(h_j)\\
    s_{i,j}^{edge} &= v_i^{de} U^{edge} v_j^{he}\\
    &= \text{DeepBiaffine}^{edge}(h_i,h_j)
    \end{align*}where $v_i^{he}$ ($v_i^{de}$) is the vector representation of $w_i$ when it is considered as a head (dependent) of a dependency relation, $s_{i,j}^{edge}$ is the score of having an arc from $w_i$ to  $w_j$, and $U^{edge}$ represents the biaffine transformation matrix for edge scores. $\text{DeepBiaffine}^{edge}$ is the combination of the FC layer followed by the biaffine transformation. Similarly, the dependency label scores are generated using two label-specific FC layers ($FC_{head}^{label}$ and $FC_{dep}^{label}$) as follows:\begin{align*}
        v_i^{hl}, v_j^{dl} &= FC_{head}^{label}(h_i), FC_{dep}^{label}(h_j)\\
    s_{i,j}^{label} &= v_i^{hl} U^{label} v_j^{dl}\\
    &= \text{DeepBiaffine}^{label}(h_i,h_j)\\
    p_{i,j}^{label} &=  \underset{0\leq l \leq |L|}{\mathrm{argmax}} \quad s_{i,j}^{label}[l]
    \end{align*}where $s_{i,j}^{label}$ is a  vector of length $|L|$ representing scores for each label type for a dependency relation where the head is $w_j$ and the dependent is $w_i$, $L$ is the set of all dependency labels, and $p_{i,j}^{label} $ is the maximum scoring label for that particular relation which will be used during the inference step.

Our parser is a simplified version of the previous graph-based parsers~\citep{qi2018universal,dozat2018simpler}. Unlike \citet{qi2018universal} we do not take into account the distance and relative ordering between two ends of each arc. Besides, both models additionally use as input; multiple pretrained word embeddings, lemma embeddings and character-level embeddings  to boost the performance of the parsers. We replace all these inputs with the BERT embeddings and casing embeddings. Our parser outperformed both parsers when trained and tested on the same settings. 
\subsubsection{Training}During training of the parser, we combine the loss obtained for both modules of the dependency parser. The loss for the edge module is calculated as the cross entropy loss over all possible heads for a given word $w_i$.\begin{align*}
    loss_i^{edge} &= \sum_{j\neq i} (  y_i^{j} * log(s_{i,j}^{edge})\\
    &+ (1- y_i^{j}) * log(1-s_{i,j}^{edge}) )\\
    loss^{edge} &=\sum_{i} loss_i^{edge}
\end{align*}where $y_i$ is the one-hot encoding of the gold label head index $y_i^{head}$ for word $w_i$,$$
y_j^i=\begin{cases}
               1 \quad \text{if  $w_j$ is the head of $w_i$.}\\
               0 \quad \text{o.w.}
            \end{cases}
$$Similarly, the loss for the label component uses cross entropy. First we get the $|L|$ long vector $s_{i,j}^{label}$ by using the gold label head index $y_i^{head}$ for a given word $w_i$ to find the head index. Then we use the same formulation above to find the loss for the label component:\begin{align*}
    y_i^{*label} &= s_{i,y_i^{head}}^{label}\\
    loss_{i}^{label} &= \sum_{l \in L}  (y\_label_i^{l} * log(y_{i}^{*label}[l])\\
    &+ (1- y\_label_i^{l}) * log(1-y_{i}^{*label}[l]) )\\
    loss^{label} &= \sum_{i} loss_{i}^{label}
    \end{align*}where $y\_label_{i}$ is the one-hot encoding of the true dependency label $y_{i}^{label}$ for $w_i$, and $y_{i}^{*label}[l]$ denotes the score for the $l^{th}$ label. Finally, we sum the two loss values to find the total loss for the dependency parser: $loss^{dep} = loss^{edge} + loss^{label}$.
\begin{figure}
    \centering
    \includegraphics[scale=0.23
    ]{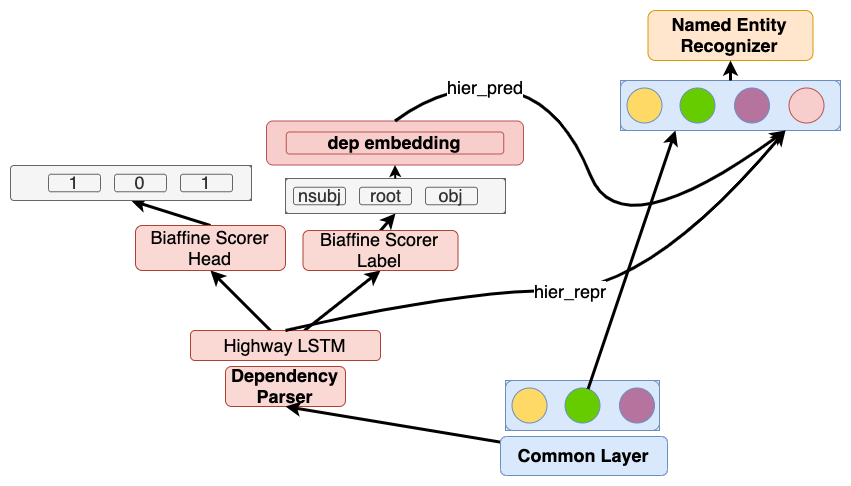}
    \caption{The detailed architecture of the hierarchical multitask setting when DEP is modeled to be the low-level task.}
    \label{detailed}
\end{figure}
\subsubsection{Inference}

During prediction mode, we first pick the highest scoring label, $p_{i,j}^{label}$, for an arc from $i$ to $j$. Then, following Qi et al.~\cite{qi2018universal}, we find the highest scoring spanning tree using Chu-Liu/Edmonds Algorithm \cite{chu1965shortest,edmonds1967optimum}. The algorithm takes as input $s_{i,j}^{edge}$ values and outputs the highest scoring unlabeled dependency tree prediction $\textbf{y}^{*edge}$.


\subsection{Named Entity Recognizer}

The NER-component (right side of Figure~\ref{fig:pyJNERDEP}) consists of three layers: bidirectional LSTM (biLSTM) layer, Fully Connected layer, and Conditional Random Fields (CRF) layer. This combination of LSTM and CRF is commonly used for the NER task~\cite{lample2016neural,ma2016end}. We extend the previous work on BiLSTM-CRF based architectures by replacing the traditional BiLSTM layer with the Highway-LSTM architecture~\cite{srivastava2015highway}. The architecture of the Highway-LSTM is identical with the DEP component.

For the hierarchical MTL framework, we consider three different architectures where the main difference is the way the task-specific components communicate. Here we only explain the settings where NER is considered as the high-level task. The settings where NER is considered as the low-level task are defined similarly. The first model, $flat$, is similar with the Multi Task Deep Neural Networks(MT-DNN) framework~\cite{liu2019multi}. In this setting, task-specific components only share the common layer shown in Figure~\ref{detailed} and the arcs from the parser to the NER component (hier\_repr and hier\_pred) are null. In the second setting, which we call $hier\_pred$ as an abbrevation for `hierarchical and prediction', the higher level component receives the embedding of the prediction made by the low level task component. First, the DEP component gets the input sentence and generates a dependency label prediction for each word. Next, we use an embedding layer to generate the vector representation of the label type. To get the label representation, we first use a simplistic assumption and pick the highest scoring label of the highest scoring arc for each word.\begin{align*}
    head_i &=  \underset{j\in \{1,...,N\}}{\mathrm{argmax}} s_{i,j}^{edge}\\
    label\_max_i &= p_{i,head_i}^{label} \\
    v^{dep}_{i} &= \text{DEP\_EMBED}(label\_max_i)\\
    o_i &= shared(w_i) = v^{bert}_{i} \oplus v^{cas}_{i} \oplus v^{pos}_{i} \\
    v_i &= o_i \oplus v^{dep}_i
\end{align*}where $p_{i,head_i}^{label}$ is the highest scoring label prediction for an arc from $w_i$ to $w_{head_i}$, $v_i$ is the input vector to the NER component, and DEP\_EMBED is the function that maps a dependency label to a fixed-size vector. We refer to this method of inputting the prediction embedding as `hard'.
Alternatively, we also used a weighted average of the embeddings, which we refer to as `soft', for each  dependency label type based on the scores generated for each of them, also employed by~\citet{Hashimoto2017AJM}. In this setting $v_i^{dep}$ is defined as \begin{align*}
v_i^{dep} &= \sum_{l\in L} \text{DEP\_EMBED}(l)   \dfrac{e^{s_{i,head_i}^{label}[l]}}{\sum_{l \in L}e^{s_{i,head_i}^{label}[l]}}\\
 &= \sum_{l\in L} \text{DEP\_EMBED}(l)  prob_i(l)
\end{align*}For the third setting, which we call `hier\_repr' to denote `hierarchical representation', we consider directly concatenating the hidden layer output generated for the DEP component (arrow denoted as `hier\_repr' on Figure~\ref{detailed}). \begin{align*}
    h^{dep}_{i} &= HLSTM^{dep} (o_1,...,o_n;i)\\
    o_i &= shared(w_i) = v^{bert}_{i} \oplus v^{cas}_{i} \oplus v^{pos}_{i} \\
    v_i &= o_i \oplus h^{dep}_i
\end{align*}where $h_{i}^{dep}$ is the final hidden layer output of the HLSTM of the DEP component for $w_i$. For the remainder of this paper, when NER is considered to be the high-level (low-level) task we refer to it as `NER\_high (low)\_pred/repr' to differentiate between different architectures. The cases for DEP is symmetric.

\subsubsection{Training}
During training, we use negative log-likehood as the objective loss function for minimization.\begin{align*}
    h_j^{ner} &= HLSTM^{ner}(v_1, ..., v_j,... v_n;j)\\
s_j &= FC^{ner} (h_j^{ner})\\
\text{\textbf{S}} &= [s_1,...,s_j,...,s_n]\\
crf\_loss &= forward\_score(\textbf{S},\textbf{T})\\
    &- path\_score(\textbf{S},\textbf{T},G)
\end{align*} where \textbf{S} denotes the scoring matrix containing scores for each label and word pair, and \textbf{T} is the transition matrix containing transition scores between each label. $forward\_score(\textbf{S},\textbf{T})$ denotes the total score of all paths and $path\_score(\textbf{S},\textbf{T},G)$ is the score of the gold label sequence. Ideally we want all probabilities to accumulate on the gold-label path so that these two scores will be identical.

\subsubsection{Inference}

We find the highest scoring label sequence for a given sentence using the Viterbi decoding algorithm~\cite{forney1973viterbi}. Formally for a given sentence $s=(w_1,...,w_n)$, the output is $\textbf{P}^* = (p_1, p_2, ... , p_n)$ which satisfies $$\textbf{P}^* = \underset{P}{\mathrm{argmax}} \quad path\_score(\textbf{S},\textbf{T},P)$$

\section{Experimental Settings}

For all the experiments, we used Tesla V100 GPU's with a single thread. The multitask model with the highest number of parameters trains at a speed of around 200 sentences/second. During the inference mode, the model can output predictions with 600 sentences/second.

\subsection{Datasets}
For both tasks, we used  the commonly used benchmarks for the Turkish language. As the dependency parsing dataset, we used the `tr-imst\_ud' dataset annotated for the `CoNLL 2018 Universal Dependencies Shared Task'~\cite{zeman-EtAl:2018:K18-2}. The dataset is annotated in CoNLL-U format and contains XPOS tags in addition to the head and dependency label for each token. Other fields that contain lemmatization, UPOS (language independent POS tags in Universal Dependency format), and morphological features are not used in our study to keep the feature-dependency at minimum. For evaluation, we report results using the two most common scoring metrics for the DEP task: Labeled Attachment Score (LAS) and Unlabeled Attachment Score (UAS). LAS considers a prediction correct only if both the head index and the dependency label are predicted correctly. For UAS the latter constraint is removed.

As the named entity recognition dataset, we used the dataset by~\citet{tur2003statistical} which contains named entity labels for Location (LOC), Organization (ORG), and Person (PER) for the NER task. The annotation is done using the IOB-2 tagging scheme where the first token of each entity has the prefix `B-' and the remaining tokens are prefixed with an `I-'. To evaluate the performance of the NER  component we use the micro F-1 score over all entities.

\subsection{Training Details}

For all the reported results of our models, we used a default batch-size of 500 words which roughly corresponds to 50 sentences on average for both datasets and define an epoch as 100 steps to have a roughly equal evaluation interval with the StanfordNLP dependency parser~\cite{dozat2018simpler}. We evaluated the performance on the validation datasets and used early stopping to stop training when we could not observe any improvements on the validation splits. See Appendix~\ref{traindets} for more details about the training procedure.

\subsubsection{Hyperparameter optimization}

In deep-learning based MTL settings there are various hyperparameters and an eager attempt to exhaustively search over the entire hyperparameter space is almost always infeasible.
To find a good hyperparameter configuration we used a bayesian optimization-based hyperparameter optimizer~\cite{bergstra2013making}. We ran the optimizer with 50 trials for each task-specific component separately. The objective function of the optimizer to minimize is the negative of the evaluation metric for each task. For each such configuration, training is done until at most 40 epochs and the maximum F-1 score attained is used. For all training settings we used learning rate patience of three epochs and early stopping with ten epochs (see Appendix Table~\ref{hypers} for the final hyperparameter configuration).

\section{Results}

In this section we evaluate the performance of each proposed setting. `NER' followed by high (low) and repr (pred) refers to the setting where NER component is considered to be the high (low) level task and the model is as described previously as $hier\_repr (pred)$. Observe that these definitions have symmetric correspondents with the DEP components as one component being low denotes the setting where the other is the high-level as we only have two tasks. For example, DEP\_low\_pred refers to the same architecture setting with NER\_high\_pred.

\begin{table}
\centering

\begin{tabular}{c|c|c}\hline
\textbf{Model}& \textbf{LAS}& \textbf{UAS}\\\hline

DEP\_only & 67.88  & 73.80 \\
DEP\_flat & 68.52  & 74.46 \\
DEP\_low\_repr & 68.05  & 73.63\\
DEP\_low\_pred& \textbf{68.87} & \textbf{74.60}  \\
DEP\_high\_repr& 68.07   & 73.86 \\
DEP\_high\_pred & 68.32  & 74.10  \\\hline
\citep{akdemir2019joint}& 50.01& 61.11 \\
\citep{dozat2018simpler}& 64.86 & 71.55 \\
\citep{che2018towards}& 66.44 & 72.25 \\\hline
\end{tabular}
\caption[Dependency Parsing results on the tr\_imst-ud-test dataset.]{Dependency Parsing results on the tr\_imst-ud-test dataset (F-1 Score (\%)). See Appendix~\ref{related} for details about each work appearing on the table.}
\label{table:depres}
\end{table}

\begin{table}[ht]
\centering

\begin{tabular}{c|c}\hline
\textbf{Model}& \textbf{F-1 Score (\%)} \\\hline

NER\_only & \textbf{93.82    } \\
NER\_flat & 92.80 \\
NER\_low\_repr &  \textbf{93.72} \\
NER\_low\_pred& 93.17  \\
NER\_high\_repr& 93.05 \\
NER\_high\_pred& 92.68  \\\hline
\cite{yeniterzi-2011-exploiting}& 88.94  \\
\cite{akdemir2019joint}& 89.21 \\
\cite{demir2014improving}& 91.96\\
\cite{cseker2012initial}& 91.94  \\
\cite{gungor2018improving}& 93.59  \\
\cite{gunecs2018turkish}& 93.69\\\hline
\end{tabular}
\caption[NER]{NER results. Our proposed model outperforms previous state-of-the-art models \cite{gunecs2018turkish,gungor2018improving} both of which require a separate morphological analyzer and a morphological disambiguator. We observed improvements in both single task setting and in a hierarchical multi-task setting over the previous best results. See Appendix~\ref{related} for details about each work appearing on the table.}
\label{table:nerres}
\end{table}

First, we trained each task-specific component separately (NER/DEP\_only). This step is necessary to verify our initial claim on replacing token-level and morphological embeddings with contextual subword embeddings. The results (see Tables~\ref{table:depres} and~\ref{table:nerres}) show that by incorporating contextual subword embeddings and using refined architectures in both components we see a significant improvement over the previously achieved state-of-the-art results on both tasks (+1.44\% for LAS and +1.55\% for UAS for the DEP task and +0.13\% F-1 for the NER task). The models achieve this by using minimal features (pos tags and casing features) compared to the competitive models. We also replaced character-level+word-level+morphological embeddings with a single contextual subword embeddings layer and still achieved better results. The observations support our initial claim about substituting morphological analysis embeddings with subword contextual ones. In addition, compared to the joint learner in the same setting~\citep{akdemir2019joint}, we obtained significant improvements for both tasks (+17.87\% for LAS and +12.69\% for UAS for the DEP task and +4.61\% F-1 for the NER task). 

Training the `flat' model where each task-specific component update the weights of the common layer resulted in a further improvement over training the DEP component separately (+0.64\% for LAS and +0.66\% for UAS). This suggests the information retained from the NER dataset about the embeddings is useful on the DEP dataset. Considering that the NER dataset used during this experiment is around 6 times larger than the DEP dataset (27,465 vs 4,660 sentences) this improvement has a straightforward explanation.

\subsection{Hierarchical MTL Results}

In the hierarchical settings, our results are inline with the observations made by~\citet{Hashimoto2017AJM}. We also observed that the low-level task benefits from the hierarchical settings (+0.35 LAS for hier\_pred). Yet the performance of the DEP component decreased for the hier\_repr setting. When DEP is considered as the high-level task we could not observe improvements over the flat model, yet the results are still better than training the DEP component separately. 

For the NER task we could not observe further improvements over the single model by using any multi-task learning setting. However, we see that three out of four hierarchical settings have higher results compared to the previously proposed flat model~\citep{liu2019multi} (+0.92\%, +0.37\%, +0.25\% and -0.12\% F-1). We further observe that NER\_low\_repr is also outperforming the previous state-of-the-art models which only focus on the NER task.

\subsection{Variations}

\begin{figure}[ht]
    \centering
    \includegraphics[scale=0.25]{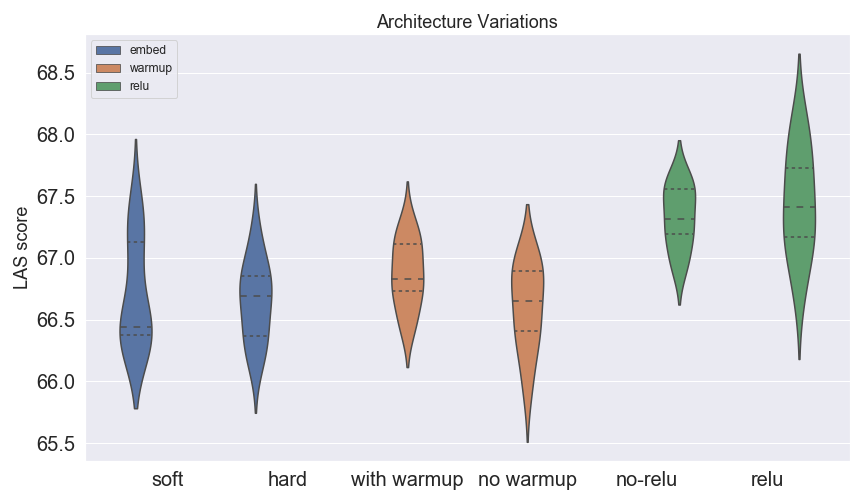}
    \caption{Performance of the DEP component for different variations. }
    \label{archvar}
\end{figure}

We also experimented with different variations of the proposed model. For each variation we fixed all other hyperparameters and trained ten models with less number of training steps. Figure~\ref{archvar} gives the distribution of the results obtained for each variation for the DEP task.

First we observed the effect of using weighted embeddings (soft) in the hierarchical settings. Especially for the setting where we use the embedding representation of the dependency prediction we expect a performance improvement over using the embedding of the prediction with highest score (hard). The best result obtained for LAS is $~68\%$ which means that almost one third of predictions either have a wrong head index or a wrong label prediction. Inputting the embedding of the highest scoring predictions in this way to the upper-level task results in updating the wrong connections on the scoring layer (deep biaffine classifiers). In addition, this damages the NER-component performance as it is informed with a misleading information about the input token. Weighted averaging does not suffer from these drawbacks. Even though the difference between the two variations are not conclusive, soft embeddings help attain higher LAS scores for some training instances. We also observed that using relu to the HLSTM outputs (compared to using no activation) and using warmup parameters bring slight improvements as well, so we kept both in the final models.

\section{Conclusion}

In this study, we proposed a hierarchical multi-task framework using subword contextual embeddings for languages with rich morphology. Our empirical results show that using subword contextual embeddings can be a good replacement for morphological analysis which is costly to obtain. By refining the task-specific components for each task and by incorporating the subword contextual embeddings we outperformed state-of-the-art models for both Dependency Parsing and Named Entity Recognition for the Turkish language. The results also show that incorporating multi-task learning brings further improvements.

To further verify our initial claim, we will report the results for other languages with rich morphology. Further gains may be possible by incorporating character-level embeddings and through extending the multi-task framework to include more tasks. To promote future research, we make all the source-code publicly available\footnote{Link to the anonymized version of our repository: https://anonymous.4open.science/r/1afb86c7-4340-4c2a-9f0b-f1145f67eb80/ (check the empty spaces in the url when copying the anonymous link manually)}.

\newpage
\bibliography{acl2020}
\bibliographystyle{acl_natbib}

\newpage
\input{appendix.tex}

\end{document}

%% file: appendix.tex
\appendix
\label{appendix}
\section{Related Work}
\label{related}
\subsection{Using morphological information for NER and DEP.}
Exploiting morphological information for NER and DEP is widely studied for the Turkish language. Yet, no previous study investigated the effect of replacing morphological analyses with subword contextual representations for NER and DEP tasks for the Turkish language before. For the NER task, \citet{yeniterzi2011exploiting} used a CRF-based model that uses morphological features and showed significant performance gains compared to a standard word-level model. \citet{cseker2012initial} also used a CRF-based model with hand crafted morphological features together with a lexicon to obtain state-of-the-art results.
\citet{demir2014improving} proposed leveraging word-level embeddings for languages with rich morphology and outperformed state-of-the-art for the Czech and Turkish languages.  Recently, \citet{gungor2018improving,gunecs2018turkish} used morphological embeddings with a BiLSTM-CRF based neural network model to obtain the current state-of-the-art results. For the DEP task, \citet{eryiugit2008dependency} showed the importance of using morphological units instead of surface forms by obtaining significant improvements. 

The DEP component proposed in our work is heavily influenced by the dependency parser of \citet{dozat2018simpler}. They use a graph-based parser and generate label and arc scores using deep biaffine transformation. \citet{che2018towards} achieved the highest overall LAS and the highest LAS for the `tr\_imst\_ud' treebank on the `CoNLL 2018 Shared Task'. Similar to our work, they make use of ELMo word-level contextual embeddings to extend the parser proposed by \citet{dozat2018simpler}. 

\subsection{Multi-task learning}
Multi-task learning is also well studied in the NLP domain. \citet{collobert2011natural} proposed a neural network based model to learn multiple tasks at once using the same level for each task. \citet{sogaard2016deep,Hashimoto2017AJM} proposed hierarchical multi-task settings to leverage obtained from low-level tasks. More related to our work, \citet{liu2019multi} also incorporated subword level contextual embeddings in a multi-task setting and obtained state-of-the-art results for various NLP tasks. Recently, \citet{akdemir2019joint} also proposed learning DEP and NER in a hierarchical multi-task setting using a BiLSTM-CRF based neural network for the NER component. The DEP component of their model is also a graph-based parser where Fully Connected layers are used to generate scores for the vector representation obtained from a BiLSTM layer.

\subsection{Contextual Embeddings for Sequence Labeling}

Using contextual embeddings at both word-level and subword level for various sequence labeling tasks is also studied previously. \citet{che2018towards} used word-level contextual embedding of ELMo~\cite{peters2018deep} for dependency parsing. \citet{liu2019multi} used subword-level contextual embeddings for various Natural Language Understanding tasks and achieved state-of-the-art results. \citet{heinzerling2019sequence} compared using contextual and non-contextual embeddings for the NER task for high-resourced and low-resourced languages.
\section{Implementation Details}

All models are implemented using Python 3.7.2 programming language and the PyTorch library version 1. The source code is publicly available at our GitHub of which we share the anonymized version for the blind-review phase  ).\footnote{https://anonymous.4open.science/r/1afb86c7-4340-4c2a-9f0b-f1145f67eb80/ (check the empty spaces in the url when copying the anonymous link manually)} In addition, we share a Docker image that contains all the requirements to run the models and replicate the results to ease the environment setup process on the repository. 

\section{Training details}
\label{traindets}
Following~\citep{dozat2018simpler}, We initialized all embeddings uniformly from the range $\left(-\sqrt{\dfrac{6}{(|L|+dim)}},+\sqrt{\dfrac{6}{(|L|+dim)}}\right)$ where $L$ denotes the number of categories for a given embedding and $dim$ is the output dimension. BERT weights are initialized using the pretrained multilingual model weights. For fine-tuning, we use the hyperparameter setting suggested by~\citet{devlin2019bert}.

One of the main challenges of MTL is the difficulty of finding the right training setting to achieve the maximal performance over all tasks. \citet{mccann2018natural} previously showed the effect of different learning strategies and the sensitivity of deep neural network (DNN) based MTL models to the changes in those settings. We follow~\citet{collobert2011natural} to randomly pick a mini-batch from the two tasks at each step. For the hierarchical models, we use a `warmup' parameter to denote the number of epochs to train the low-level task component before starting the training on the high-level task.  This allows the low-level component to generate more meaningful representations~\cite{Hashimoto2017AJM}. 

For early stopping, we used a hyperparameter to determine the patience. For the multi-task settings, if one of the tasks exceeds the patience parameter we stopped the training for both components. 
\subsection{BERT}

We used the Transformers library~\citep{wolf2019transformers} for fine-tuning the BERT model. The library provides deploying the multilingual BERT model that can be used directly for sentence-level tasks. In addition, the library provides a multilingual tokenizer to obtain the multilingual BERT tokens for a given input sentence. In our study, we refrained from using our own subword vocabulary for the Turkish language as we would like to keep the model as universal as possible and as we plan to experiment with other languages with rich morphology in future.

BERT uses multi-layer bidirectional transformers to obtain contextual subword embeddings. Bidirectional transformers make use of the attention mechanism~\citep{vaswani2017attention} to attend to both the previous and subsequent input information. Attention mechanism allows to control the information obtained from any surrounding context. For pre-training, they define two unsupervised learning tasks: Masked word prediction and next sentence prediction. For the masked word prediction task, the model is trained to minimize the loss for predicting a masked word chosen arbitrarily from a given input sequence. For the latter task, the model tries to predict whether the next sentence is a random sentence or is the actual next sentence on the training dataset.

\subsubsection{BERT tokenization mapping.}
\label{berttoks}
As mentioned previously, we first used the BERT multilingual tokenizer provided by~\citet{wolf2019transformers} to obtain the BERT tokens for a given input. In order to increase the tokenization accuracy we could define our own subword vocabulary. Yet, this increases the amount of manual human intervention required for each language and makes our model incompatible with the original multilingual BERT pretrained model. As we are planning to apply this proposed architecture in a multilingual setting we refrained from experimenting with our own subword vocabulary. We take the average of all subword embeddings for a given actual token. We observed that using only the representation of the first subword token as in~\cite{heinzerling2019sequence} or summing performed poorly compared to our approach in this context. In order to map between BERT sub-word tokens and real input tokens we need to apply a two-step procedure in order to be able to use batching. For sequences of varying lengths such as sentences, we have input vectors of different lengths. When batching is used all input vectors for a given batch should be of same size in order to form a tensor. To this end, we include an additional token in the vocabulary `[PAD]' to pad all vectors up to the length of the maximum vector of a given batch. To use BERT sub-word tokens we first apply padding to the real input tokens in a given batch to match their lengths. Recall that sub-word tokens of BERT are of varying length unlike fixed-length n-grams. This causes the sequence lengths to change after tokenizing the initial sentences with the BERT tokenizer. Thus, we apply a second padding to the padded sequences to get a proper batch input for BERT. To give an example, consider the following simple batch consisting of two sentences with lengths 1 and 2: `Geliyoruz' and `Geldi mi' which can be translated as , `we are coming' and `did he/she come', respectively. In order to give these two input sequences to the task specific layers, we pad the first sentence to match length 2: `Geliyoruz [PAD]'. Next BERT tokenizer outputs the following tokens: ['gel', '\#\#iy', '\#\#or', '\#\#uz', '[PAD]'] and ['gel', '\#\#di', 'mi']. Now the lengths of each sequence became 5 and 3. Next we need to do a second batching to match the BERT lengths : ['gel', '\#\#iy', '\#\#or', '\#\#uz', '[PAD]'] and ['gel', '\#\#di', 'mi', [PAD], [PAD]]. After getting the output vector representations from BERT, we need to remove only the paddings added for BERT to obtain the original padded sequences. To get the vector representation for a given real token we do the following: First we get the mean of the final four layers of BERT hidden layers. Next we take the mean of these mean values of each sub-word token for a given original token. For instance, for the original token  $w_j$ = `geliyoruz', we first get the mean of all four final hidden outputs for each sub-word token `gel', `\#\#iy', `\#\#or' and `\#\#uz' separately. Next we take the mean of these 4 vectors to get final representation for geliyoruz.

\begin{align*}
    V &= \{i: v_i \in \text{BERT\_tokenize}(w_j)\}\\
    v_i^{mean} &= 1/4 \sum_{l=1}^4 \text{BERT}(v_i)[-l]\quad \forall v_i \in V\\
    bert_j &= 1/|V| \sum_{i=1}^{|V|} v_i^{mean} \\ 
\end{align*}
where $bert_j$ represents the final output of the BERT module for a given word $w_j$ and $[-l]$ denotes the $l^{th}$ last element of a vector, following the Python conventions.

\subsection{Highway LSTM}
\label{hlstm}
Following the success of the dependency parser implemented in~\cite{qi2018universal} exploiting Highway LSTM's, we incorporated Highway LSTM's (HLSTM) into both task-specific components. Below we give the formulation of the HLSTM proposed by~\cite{srivastava2015highway}. One of the major challenges in training networks is the information flow between units. LSTM's are shown to be a good way of overcoming the problem of conditioning neural networks on distant inputs. By using gating and forget mechanisms LSTM cells allow networks to learn the information transfer between sequences of inputs. We can think of this information flow as the horizontal flow. Another type of information flow is the information flow between consecutive layers of LSTM's which we can think of as the vertical flow. Conventionally we input the hidden representations obtained from the previous LSTM layer for each token to the next LSTM layer. Yet, the difficulty of training deep neural networks with large number of consecutive LSTM layers is shown in previous studies~\cite{raiko2012deep,graves2013generating}. \cite{srivastava2015highway} propose a similar mechanism with LSTM's to control the information flow between stacks of LSTM layers. Conventionally, for $L$ layers the feed-forward step for a sequence of $n$ inputs can be shown as follows:
\begin{align*}
h_t^{1} &= LSTM^{1}(x_1,...,x_t,...,x_n;t)\\
h_t^{2} &= LSTM^{2}(h_1^{1},...,x_t^{1},...,x_n^{1};t)\\
&...  \\
h_t^{i} &= LSTM^{i}(h_1^{i-1},...,x_t^{i-1},...,x_n^{i-1};t)\\
&...\\
y_t &= LSTM^{L}(h_1^{i-1},...,x_t^{i-1},...,x_n^{i-1};t)\\
\end{align*}

Instead, Highway LSTM's control the vertical information flow using a gating function with learnable parameters. Additional gating functions are defined such that for $\textbf{x}=(x_1,...,x_n)$

\begin{align*}
h_t^{1} &= T^{1}(\textbf{x};t)*LSTM^{1}(\textbf{x};t)+ C^{1}(\textbf{x};t)\\
h_t^{i} &= T^{i}(\textbf{h}^{i-1};t)*LSTM^{i}(\textbf{h}^{i-1};t)\\
&+ C^{i}(\textbf{h}^{i-1};t)\\
\end{align*}

where $T$ and $C$ control the information flow from the previous LSTM-layer.In our experiments we always set $C^{i}(\textbf{x};t) = 1 - T^{i}(\textbf{x};t) $ for simplicity following~\cite{srivastava2015highway} and set $T = \sigma(\textbf{W}x + \textbf{b})$ where $\textbf{W}$ is the transformation matrix, $\textbf{b}$ is the bias vector and $\textbf{x}$ is the input to the current layer. Using $\sigma$ assures  $C+T=1$ for all inputs. Observe that for this setting, if we set $T^{i}(\textbf{h}^{i-1};t) = 1$ and $0$ the output will be equivalent to $LSTM^{i}(\textbf{h}^{i-1};t)$ and $\textbf{h}^{i-1}$ respectively. 

\begin{table}[ht]

\begin{center}
\begin{tabular}{c  c}\hline
 \multicolumn{2}{c} {\textbf{Hidden sizes}}\\\hline
casing  emb.& 32\\
pos emb.& 64 \\
dep emb.& 128 \\
ner emb.& 128 \\
bert& 768  \\
lstm hidden units& 2*400 \\
biaffine inner & 400 \\
lstm layers& 3 \\
 \multicolumn{2}{c} {\textbf{Dropout}}\\\hline
ner lstm dropout & 0.43 \\
dep lstm dropout& 0.34 \\
bert dropout& 0.5 \\
embedding dropout& 0.4 \\
\multicolumn{2}{c} {\textbf{Training}}\\\hline
optimizer& AdamW\\
learning rate (lr)& 0.004\\
lr decay& 0.1286\\
lr patience& 3 epochs\\
weight decay& 0.001\\
early stop& 10 epochs\\
warmup& 5\\
batchsize & 500\\\hline
\end{tabular}
\caption[Parameters for the model]{Hyperparameters of the model and the final configuration.}
\label{hypers}
\end{center}
\end{table}

\section{Hyperparameters}

As mentioned on the real text, we used a bayesian-optimization based hyperparameter optimizer~\cite{bergstra2013making} to get a good hyperparameter configuration for our multi-task settting. The list of the hyperparameters together with the final values are given in Table~\ref{hypers}. `pos emb.' refers to the size of the vector to represent each POS tag category. `casing emb.', `dep emb.' and `ner emb.' are defined similarly. Bidirectional LSTM units are used inside HLSTM architecture. Thus, 2*400 on Table~\ref{hypers} denotes the concatenation of forward and backward representations (400+400).

\section{Rare and unknown word graphs}
\label{graphs}
To find the word counts in each dataset, we fixed a dataset size of 30,000 words. Next we uniformly sample sentences from each dataset until we reach the desired dataset size. This is repeated for ten times and the average counts are reported on the figure.

Unknown words are defined as the words in the test set that are not seen previously on the training set. Similarly, to find the number of unknown words we uniformly sample sentences from the test sets until we reach 1000 words. Then for each $n \in \{5000,10000,15000,20000,25000,30000\}$, where $n$ is the number of words, we sampled  sentences from the training set for ten times. We report the average of unknown words for each training set size on the figure.

%% file: acl2020.bbl
\begin{thebibliography}{36}
\expandafter\ifx\csname natexlab\endcsname\relax\def\natexlab#1{#1}\fi

\bibitem[{Akdemir and G{\"u}ng{\"o}r(2019)}]{akdemir2019joint}
Arda Akdemir and Tunga G{\"u}ng{\"o}r. 2019.
\newblock Joint learning of named entity recognition and dependency parsing
  using separate datasets.
\newblock \emph{Computaci{\'o}n y Sistemas}, 23(3).

\bibitem[{Bergstra et~al.(2013)Bergstra, Yamins, and Cox}]{bergstra2013making}
James Bergstra, Daniel Yamins, and David~Daniel Cox. 2013.
\newblock Making a science of model search: Hyperparameter optimization in
  hundreds of dimensions for vision architectures.
\newblock In \emph{Proceedings of the 30th International Conference on Machine
  Learning (JMLR)}. Jmlr.

\bibitem[{Bojanowski et~al.(2017)Bojanowski, Grave, Joulin, and
  Mikolov}]{bojanowski2017enriching}
Piotr Bojanowski, Edouard Grave, Armand Joulin, and Tomas Mikolov. 2017.
\newblock Enriching word vectors with subword information.
\newblock \emph{Transactions of the Association for Computational Linguistics},
  5:135--146.

\bibitem[{Che et~al.(2018)Che, Liu, Wang, Zheng, and Liu}]{che2018towards}
Wanxiang Che, Yijia Liu, Yuxuan Wang, Bo~Zheng, and Ting Liu. 2018.
\newblock Towards better ud parsing: Deep contextualized word embeddings,
  ensemble, and treebank concatenation.
\newblock \emph{CoNLL 2018}, page~55.

\bibitem[{Chu(1965)}]{chu1965shortest}
Yoeng-Jin Chu. 1965.
\newblock On the shortest arborescence of a directed graph.
\newblock \emph{Scientia Sinica}, 14:1396--1400.

\bibitem[{Collobert et~al.(2011)Collobert, Weston, Bottou, Karlen, Kavukcuoglu,
  and Kuksa}]{collobert2011natural}
Ronan Collobert, Jason Weston, L{\'e}on Bottou, Michael Karlen, Koray
  Kavukcuoglu, and Pavel Kuksa. 2011.
\newblock Natural language processing (almost) from scratch.
\newblock \emph{Journal of Machine Learning Research}, 12(Aug):2493--2537.

\bibitem[{Demir and Ozgur(2014)}]{demir2014improving}
Hakan Demir and Arzucan Ozgur. 2014.
\newblock Improving named entity recognition for morphologically rich languages
  using word embeddings.
\newblock In \emph{ICMLA}, pages 117--122.

\bibitem[{Devlin et~al.(2019)Devlin, Chang, Lee, and
  Toutanova}]{devlin2019bert}
Jacob Devlin, Ming-Wei Chang, Kenton Lee, and Kristina Toutanova. 2019.
\newblock Bert: Pre-training of deep bidirectional transformers for language
  understanding.
\newblock In \emph{Proceedings of the 2019 Conference of the North American
  Chapter of the Association for Computational Linguistics: Human Language
  Technologies, Volume 1 (Long and Short Papers)}, pages 4171--4186.

\bibitem[{Dozat and Manning(2018)}]{dozat2018simpler}
Timothy Dozat and Christopher~D Manning. 2018.
\newblock Simpler but more accurate semantic dependency parsing.
\newblock In \emph{Proceedings of the 56th Annual Meeting of the Association
  for Computational Linguistics (Volume 2: Short Papers)}, pages 484--490.

\bibitem[{Edmonds(1967)}]{edmonds1967optimum}
Jack Edmonds. 1967.
\newblock Optimum branchings.
\newblock \emph{Journal of Research of the national Bureau of Standards B},
  71(4):233--240.

\bibitem[{Eryi{\u{g}}it et~al.(2008)Eryi{\u{g}}it, Nivre, and
  Oflazer}]{eryiugit2008dependency}
G{\"u}l{\c{s}}en Eryi{\u{g}}it, Joakim Nivre, and Kemal Oflazer. 2008.
\newblock Dependency parsing of turkish.
\newblock \emph{Computational Linguistics}, 34(3):357--389.

\bibitem[{Forney(1973)}]{forney1973viterbi}
G~David Forney. 1973.
\newblock The viterbi algorithm.
\newblock \emph{Proceedings of the IEEE}, 61(3):268--278.

\bibitem[{Graves(2013)}]{graves2013generating}
Alex Graves. 2013.
\newblock Generating sequences with recurrent neural networks.
\newblock \emph{arXiv preprint arXiv:1308.0850}.

\bibitem[{G{\"u}ne{\c{s}} and Tantu{\u{G}}(2018)}]{gunecs2018turkish}
Asim G{\"u}ne{\c{s}} and A~C{\"u}neyd Tantu{\u{G}}. 2018.
\newblock Turkish named entity recognition with deep learning.
\newblock In \emph{2018 26th Signal Processing and Communications Applications
  Conference (SIU)}, pages 1--4. IEEE.

\bibitem[{Gungor et~al.(2018)Gungor, Uskudarli, and
  Gungor}]{gungor2018improving}
Onur Gungor, Suzan Uskudarli, and Tunga Gungor. 2018.
\newblock Improving named entity recognition by jointly learning to
  disambiguate morphological tags.
\newblock In \emph{Proceedings of the 27th International Conference on
  Computational Linguistics}, pages 2082--2092.

\bibitem[{Hashimoto et~al.(2017)Hashimoto, Xiong, Tsuruoka, and
  Socher}]{Hashimoto2017AJM}
Kazuma Hashimoto, Caiming Xiong, Yoshimasa Tsuruoka, and Richard Socher. 2017.
\newblock A joint many-task model: Growing a neural network for multiple nlp
  tasks.
\newblock In \emph{EMNLP}.

\bibitem[{Heinzerling and Strube(2019)}]{heinzerling2019sequence}
Benjamin Heinzerling and Michael Strube. 2019.
\newblock Sequence tagging with contextual and non-contextual subword
  representations: A multilingual evaluation.
\newblock \emph{arXiv preprint arXiv:1906.01569}.

\bibitem[{Hochreiter and Schmidhuber(1997)}]{hochreiter1997long}
Sepp Hochreiter and J{\"u}rgen Schmidhuber. 1997.
\newblock Long short-term memory.
\newblock \emph{Neural computation}, 9(8):1735--1780.

\bibitem[{Lample et~al.(2016)Lample, Ballesteros, Subramanian, Kawakami, and
  Dyer}]{lample2016neural}
Guillaume Lample, Miguel Ballesteros, Sandeep Subramanian, Kazuya Kawakami, and
  Chris Dyer. 2016.
\newblock Neural architectures for named entity recognition.
\newblock In \emph{Proceedings of the 2016 Conference of the North American
  Chapter of the Association for Computational Linguistics: Human Language
  Technologies}, pages 260--270.

\bibitem[{Liu et~al.(2019)Liu, He, Chen, and Gao}]{liu2019multi}
Xiaodong Liu, Pengcheng He, Weizhu Chen, and Jianfeng Gao. 2019.
\newblock Multi-task deep neural networks for natural language understanding.
\newblock \emph{arXiv preprint arXiv:1901.11504}.

\bibitem[{Luong et~al.(2013)Luong, Socher, and Manning}]{luong2013better}
Thang Luong, Richard Socher, and Christopher Manning. 2013.
\newblock Better word representations with recursive neural networks for
  morphology.
\newblock In \emph{Proceedings of the Seventeenth Conference on Computational
  Natural Language Learning}, pages 104--113.

\bibitem[{Ma and Hovy(2016)}]{ma2016end}
Xuezhe Ma and Eduard Hovy. 2016.
\newblock End-to-end sequence labeling via bi-directional lstm-cnns-crf.
\newblock In \emph{Proceedings of the 54th Annual Meeting of the Association
  for Computational Linguistics (Volume 1: Long Papers)}, pages 1064--1074.

\bibitem[{McCann et~al.(2018)McCann, Keskar, Xiong, and
  Socher}]{mccann2018natural}
Bryan McCann, Nitish~Shirish Keskar, Caiming Xiong, and Richard Socher. 2018.
\newblock The natural language decathlon: Multitask learning as question
  answering.
\newblock \emph{arXiv preprint arXiv:1806.08730}.

\bibitem[{Mikolov et~al.(2013)Mikolov, Sutskever, Chen, Corrado, and
  Dean}]{mikolov2013distributed}
Tomas Mikolov, Ilya Sutskever, Kai Chen, Greg~S Corrado, and Jeff Dean. 2013.
\newblock Distributed representations of words and phrases and their
  compositionality.
\newblock In \emph{Advances in neural information processing systems}, pages
  3111--3119.

\bibitem[{Nadeau and Sekine(2007)}]{nadeau2007survey}
David Nadeau and Satoshi Sekine. 2007.
\newblock A survey of named entity recognition and classification.
\newblock \emph{Lingvisticae Investigationes}, 30(1):3--26.

\bibitem[{Peters et~al.(2018)Peters, Neumann, Iyyer, Gardner, Clark, Lee, and
  Zettlemoyer}]{peters2018deep}
Matthew Peters, Mark Neumann, Mohit Iyyer, Matt Gardner, Christopher Clark,
  Kenton Lee, and Luke Zettlemoyer. 2018.
\newblock Deep contextualized word representations.
\newblock In \emph{Proceedings of the 2018 Conference of the North American
  Chapter of the Association for Computational Linguistics: Human Language
  Technologies, Volume 1 (Long Papers)}, pages 2227--2237.

\bibitem[{Qi et~al.(2018)Qi, Dozat, Zhang, and Manning}]{qi2018universal}
Peng Qi, Timothy Dozat, Yuhao Zhang, and Christopher~D Manning. 2018.
\newblock Universal dependency parsing from scratch.
\newblock \emph{Proceedings of the CoNLL 2018 Shared Task: Multilingual Parsing
  from Raw Text to Universal Dependencies}, pages 160--170.

\bibitem[{Raiko et~al.(2012)Raiko, Valpola, and LeCun}]{raiko2012deep}
Tapani Raiko, Harri Valpola, and Yann LeCun. 2012.
\newblock Deep learning made easier by linear transformations in perceptrons.
\newblock In \emph{Artificial intelligence and statistics}, pages 924--932.

\bibitem[{{\c{S}}eker and Eryi{\u{g}}it(2012)}]{cseker2012initial}
G{\"o}khan~Ak{\i}n {\c{S}}eker and G{\"u}l{\c{s}}en Eryi{\u{g}}it. 2012.
\newblock Initial explorations on using crfs for turkish named entity
  recognition.
\newblock \emph{Proceedings of COLING 2012}, pages 2459--2474.

\bibitem[{S{\o}gaard and Goldberg(2016)}]{sogaard2016deep}
Anders S{\o}gaard and Yoav Goldberg. 2016.
\newblock Deep multi-task learning with low level tasks supervised at lower
  layers.
\newblock In \emph{Proceedings of the 54th Annual Meeting of the Association
  for Computational Linguistics (Volume 2: Short Papers)}, volume~2, pages
  231--235.

\bibitem[{Srivastava et~al.(2015)Srivastava, Greff, and
  Schmidhuber}]{srivastava2015highway}
Rupesh~K Srivastava, Klaus Greff, and J{\"u}rgen Schmidhuber. 2015.
\newblock Training very deep networks.
\newblock In \emph{Advances in neural information processing systems}, pages
  2377--2385.

\bibitem[{T{\"u}r et~al.(2003)T{\"u}r, Hakkani-T{\"u}r, and
  Oflazer}]{tur2003statistical}
G{\"o}khan T{\"u}r, Dilek Hakkani-T{\"u}r, and Kemal Oflazer. 2003.
\newblock A statistical information extraction system for turkish.
\newblock \emph{Natural Language Engineering}, 9(2):181--210.

\bibitem[{Vaswani et~al.(2017)Vaswani, Shazeer, Parmar, Uszkoreit, Jones,
  Gomez, Kaiser, and Polosukhin}]{vaswani2017attention}
Ashish Vaswani, Noam Shazeer, Niki Parmar, Jakob Uszkoreit, Llion Jones,
  Aidan~N Gomez, {\L}ukasz Kaiser, and Illia Polosukhin. 2017.
\newblock Attention is all you need.
\newblock In \emph{Advances in neural information processing systems}, pages
  5998--6008.

\bibitem[{Wolf et~al.(2019)Wolf, Debut, Sanh, Chaumond, Delangue, Moi, Cistac,
  Rault, Louf, Funtowicz et~al.}]{wolf2019transformers}
Thomas Wolf, Lysandre Debut, Victor Sanh, Julien Chaumond, Clement Delangue,
  Anthony Moi, Pierric Cistac, Tim Rault, R{\'e}mi Louf, Morgan Funtowicz,
  et~al. 2019.
\newblock Transformers: State-of-the-art natural language processing.
\newblock \emph{arXiv preprint arXiv:1910.03771}.

\bibitem[{Yeniterzi(2011)}]{yeniterzi2011exploiting}
Reyyan Yeniterzi. 2011.
\newblock Exploiting morphology in turkish named entity recognition system.
\newblock In \emph{Proceedings of the ACL 2011 Student Session}, pages
  105--110. Association for Computational Linguistics.

\bibitem[{Zeman et~al.(2018)Zeman, Haji{\v{c}}, Popel, Potthast, Straka,
  Ginter, Nivre, and Petrov}]{zeman-EtAl:2018:K18-2}
Daniel Zeman, Jan Haji{\v{c}}, Martin Popel, Martin Potthast, Milan Straka,
  Filip Ginter, Joakim Nivre, and Slav Petrov. 2018.
\newblock {CoNLL} 2018 shared task: Multilingual parsing from raw text to
  universal dependencies.
\newblock In \emph{Proceedings of the {CoNLL} 2018 Shared Task: Multilingual
  Parsing from Raw Text to Universal Dependencies}, pages 1--21, Brussels,
  Belgium. Association for Computational Linguistics.

\end{thebibliography}
